\documentclass[a4paper]{cas-sc}
\usepackage[numbers]{natbib}
\usepackage{graphicx}
\usepackage{booktabs}
\usepackage[ruled, vlined]{algorithm2e}
\usepackage{array}
\usepackage{times}
\usepackage{url}
\usepackage{caption}
\usepackage{multirow} 
\usepackage{amsthm, amsmath, amssymb}
\usepackage{microtype}
\def\tsc#1{\csdef{#1}{\textsc{\lowercase{#1}}\xspace}}
\tsc{WGM}
\tsc{QE}

\begin{document}
\let\WriteBookmarks\relax
\def\floatpagepagefraction{1}
\def\textpagefraction{. 001}

\title [mode = title]{CMD-HAR: Cross-Modal Disentanglement for Wearable Human Activity Recognition} 
\shorttitle{CMD-HAR}
\shortauthors{Li el al.}
\author{Ying Yu}
\author{Siyao Li}
\author{Yixuan Jiang}
\author{Hang Xiao}
\author{Jingxi Long}
\author{Haotian Tang}
\author{Hanyu Liu*}
\author{Chao Li*}

\begin{abstract}
Human Activity Recognition (HAR) is a fundamental technology for numerous human - centered intelligent applications. Although deep learning methods have been utilized to accelerate feature extraction, issues such as multimodal data mixing, activity heterogeneity, and complex model deployment remain largely unresolved. The aim of this paper is to address issues such as multimodal data mixing, activity heterogeneity, and complex model deployment in sensor-based human activity recognition. We propose a cross-modal spatiotemporal decomposition strategy to tackle the problem of the mixed distribution of sensor data. Key discriminative features of activities are captured through cross-modal spatio-temporal disentangled representation, and gradient modulation is combined to alleviate data heterogeneity. In addition, a wearable deployment simulation system is constructed. We conducted experiments on a large number of public datasets, demonstrating the effectiveness of the model.

\end{abstract}


%
%
%
%
%

\maketitle

\section{Introduction}

Human Activity Recognition (HAR) aims to identify the types of activities performed by humans based on information received from different devices, such as cameras and motion sensors \cite{yang2022cross}. HAR has been widely used in various real-world applications, especially in healthcare \cite{zhou2020deep} and exercise assessment \cite{zhou2023swarm}. Based on data type, HAR systems can be split into two major categories: video-based HAR and sensor-based HAR. 
However, video-based systems recognize behaviors through captured images or videos, facing societal and technical challenges, such as privacy concerns, dependence on environmental lighting, resolution constraints, and the high cost and complexity associated with video processing algorithms. In contrast, sensor-based HAR has been widely used because of their ease of installation, and non-intrusiveness \cite{essa2023temporal}. Until now, mobile devices have brought about a new way of life and have become an indispensable part of people’s daily lives \cite{yin2024systematic}.



Especially in practical application scenarios, human activities in the real world have a high degree of diversity and variability. Existing deep learning techniques have exposed significant limitations in dealing with such complex situations. The following are several key issues that are gradually emerging in current HAR research: \textbf{1) Difficult Problem of Multimodal Complex Activity Recognition.} Complex activities are composed of simple sub - activities with high - level semantic information. The fine - grained activity features of tomographic sensors increase the difficulty of recognition. The data structures and feature expressions of different types of sensors vary greatly, resulting in the heterogeneous and inconsistent distribution of multimodal data with non-independent and non-identical characteristics. This not only makes it difficult to establish a unified feature space, but also increases the complexity of feature fusion, restricts the generalization ability of the model, and makes it difficult for the model to maintain stable performance under different scenarios and datasets. \textbf{2) Challenge of Activity Heterogeneity.} For the same type of activity, the sensor data generated by different individuals often vary greatly. However, most current studies overlook the differences in activity performance between these individuals. In fact, due to individual characteristics such as gender, habits, and physical strength, the sensor records of the same activity may be different. Early deep learning models could handle some problems, but they are no longer sufficient for complex activity recognition tasks. How to accurately extract key discriminative information from complex and diverse features and effectively reduce the interference of invalid features on the recognition process is one of the core tasks in current HAR research. \textbf{3) Practical Feasibility of Complex Models.} In recent years, HAR research has adopted high - performance device testing, which reflects the potential of the technology. However, the increasing difficulty of recognition tasks has led to an increase in the complexity of the methods. For example, compared with professional high - performance testing equipment, the computing resources of actual wearable devices are extremely limited and cannot support the calculation of complex large - scale models. At the same time, to ensure the durability of the device, strict power consumption control is required, which means that the HAR model needs to optimize the algorithm during operation to reduce energy consumption.


This paper, therefore, makes three main contributions: 
\begin{itemize}

\item Firstly, the spatiotemporal disentanglement mechanism between modes addresses the heterogeneity of activities, ensuring that key patterns across time and space are consistently captured.

\item Secondly, gradient modulation between modalities solves multimodal heterogeneity by dynamically adjusting the influence of each modality based on its quality and reliability, ensuring balanced contributions during the learning process.

\item Lastly, we develop a wearable deployment simulation system using Raspberry Pi 5 devices, analyzing the model's complexity and inference latency to validate its practical feasibility.
\end{itemize}

\begin{figure}[ht]
    \centering
    \includegraphics[width=0.8\linewidth]{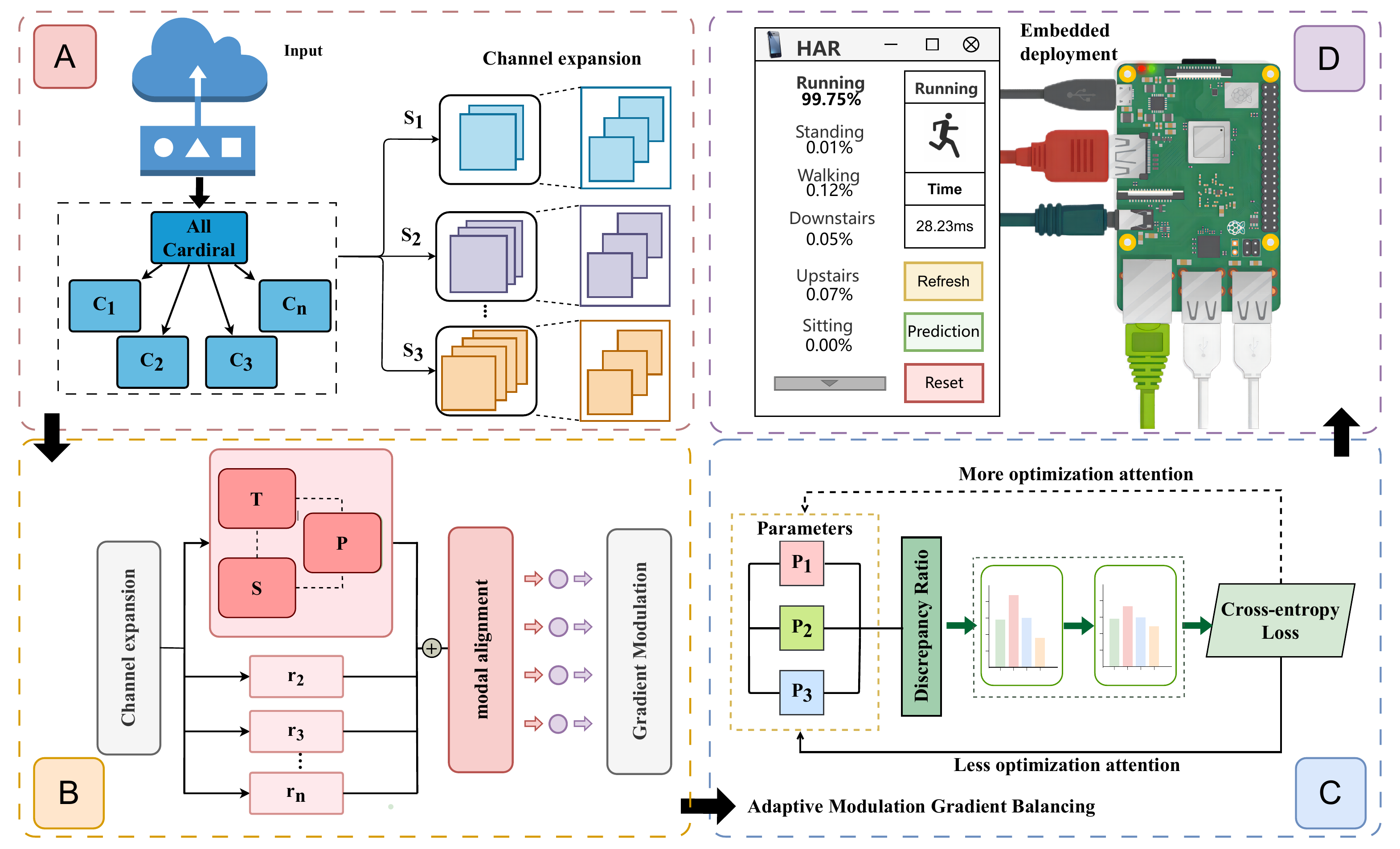}
    \caption{Framework of our CMD-HAR method, including (A) Channel expansion, (B) Spatiotemporal de entanglement module, (C) Gradient modulation, (D) Embedded deployment}
    \label{framework}
\end{figure}
\vspace{-1em}
\section{Related Works}

\subsection{Feature Extraction of Complex Activities}

Prior to the advent of deep learning, HAR systems were predominantly based on manual feature engineering. The spatiotemporal interest point (STIP) method proposed by Laptev \cite{1238378} in 2005 exemplifies this approach, utilizing spatiotemporal and geometric information for feature representation. However, such methods required extensive domain expertise and demonstrated limited efficacy in complex activity recognition. Subsequent advances introduced data-driven approaches, notably Simonyan and Zisserman's two-stream CNN (2D-CNN) \cite{DBLP:journals/corr/SimonyanZ14}, which fused spatial features from single frames with temporal features from optical flow maps. Although effective for temporal modeling, 2D-CNNs exhibited limitations in capturing long-range dependencies. This led to innovations such as P-CNN \cite{7410725}, which incorporated human body part trajectories to better model complex motions. 

\vspace{-1em}
\subsection{Disentangled Representation Learning}

Disentangled representation learning seeks to isolate key explanatory factors in data generation while preserving the interpretability of the model, a critical requirement for HAR systems. The theoretical foundation was established by Bengio et al. \cite{10.1109/TPAMI.2013.50}, who posited that disentangled representations enable better generalization through separation of latent factors. Practical implementations emerged with Higgins' $\beta$-VAE \cite{Higgins2016betaVAELB}, which introduced a tunable hyperparameter to balance reconstruction accuracy and factor disentanglement. Kim and Mnih's FactorVAE \cite{pmlr-v80-kim18b} further improved disentanglement efficacy through dedicated regularization terms. 

In HAR, the use of disentangling can strengthen the recognition of key feature information, mitigate gradient explosion issues, enable deeper model architectures, and improve the recognition performance. Cheng et al. \cite{10122911}proposed a protohar to effectively disentangle representations and classifiers in heterogeneous federated learning environments, using global prototypes to correct the representation of activity characteristics, avoiding excessive drift of local models in personalized training, and improving the personalization and convergence speed of models. Yang et al. \cite{10440498} proposed the feature-disconnecting activity recognition network (fdarn) in the field of cm-fhar. Through decentralized optimization and spherical modal identification loss, the model can not only make good generalization between different clients by using unknowable modal features, but also capture the modal specific discrimination features of each client. 
\vspace{-1em}
\subsection{Modal Heterogeneity}

Current approaches to modal heterogeneity face three principal challenges: 1) Data imputation techniques such as GANs, which have been employed to address issues like inconsistent missing data patterns in HAR, rely on high-quality training data and exhibit limited generalizability \cite{hussein2022robust}. 2) Conventional grouping strategies restrict cross-modal knowledge transfer \cite{palechor2022semi}. 3) Existing knowledge-sharing paradigms fail to fully integrate multi-scale features \cite{guo2025multi}. Gradient modulation methods provide a promising solution to address modal heterogeneity in HAR, significantly enhancing multimodal information fusion. Peng et al. \cite{Peng2022BalancedML} proposed the dynamic gradient modulation method, which adjusts modal contributions during optimization, though restricted to late fusion architectures. Li et al. \cite{li2023boosting} explored an adaptive gradient modulation (AGM) method based on Shapley values, which is applicable to complex fusion strategies and introduces a novel modality competition intensity metric to reveal the underlying mechanisms of modality competition. Guo et al. \cite{guo2024classifier} introduced the classifier-guided gradient modulation (CGGM) method, which evaluates modality utility using lightweight modal classifiers, integrating both gradient magnitude and direction. Compared to existing methods, CGGM is not restricted by loss functions, optimizers, or the number of modalities, but the introduction of additional classifiers increases computational overhead.

\section{Methodology}

Our proposed CMD-HAR is an innovative disentanglement method aimed at identifying heterogeneous multimodal complex activities. As shown in Figure \ref{framework}, detailed and specific conditions and mechanisms are explained. Specifically, the model first employs Channel Expansion modules to group the input data, thereby adapting to the characteristics of multi-sensor inputs. Subsequently, by designing independent disentanglement of the temporal and spatial dimensions, the accuracy and robustness of feature extraction are improved. In addition, to reduce the heterogeneity of data, we introduce a gradient modulation strategy to achieve efficient feature integration. Finally, we design a lightweight deployment scheme suitable for embedded devices. The specific structural description is as follows.

\subsection{Driven Feature Grouping}

To address the heterogeneity and feature distribution differences of multi-sensor inputs, we propose a grouping mechanism based on cardinality characteristics, implemented through Channel Expansion modules to enhance feature representation.
Firstly, the input feature channels are divided into multiple Cardinality Groups, each containing $C/K$ channels, where $K$ represents the number of groups, and $C$ represents the total number of feature channels. Within each group, the channels are further divided according to the Radix (the number of splits), forming sub-groups. $R$ represents the number of splits per group, and the total number of sub-groups is $G = K \times R$. The combinatorial representation for the $k$-th cardinality group is formulated as: ${X}^{k}=\sum_{j=R(k-1)+1}^{Rk}\mathbf{X}_{j},$
where${X}^{k}\in\mathbb{R}^{S\times C/K},k\in1,2,\ldots K$. This parametric grouping strategy enables adaptive sensor fusion while preserving modality-specific patterns.

\subsection{Channel Expansion}

Prior to spatio-temporal disentanglement, we deployed a radix-aware spatial attention mechanism to achieve cross-modal alignment. This mechanism achieves alignment and standardization of different modal features by assigning dynamic weights and uniformly mapping the expanded number of channels to a fixed dimension. The cardinality group represents the weighted fusion of $\mathbf{f}^{k}\in\mathbb{R}^{S\times C/K}$
using soft channel attention for aggregation, where each feature map channel is generated using weighted combinations in splitting. Specifically, the c-th channel is calculated as follows:
$\mathbf{f}_{c}^{k}=\sum_{i=1}^{R}a_{i}^{k}(c)\mathbf{X}_{R(k-1)+i}$. The weight allocation is shown by the following formula: 
 \begin{equation}
 a_{i}^{k}(c) = 
\begin{cases}
\frac{\exp(\mathcal{G}_{i}^{c}(s^{k}))}{\sum_{j=1}^{R}\exp(\mathcal{G}_{j}^{c}(s^{k}))} & R > 1, \\
\sigma(\mathcal{G}_{i}^{c}(s^{k})) & R = 1
\end{cases}
\end{equation}
where $\mathcal{G}(\cdot)$ denotes the gating function, $\sigma(\cdot)$ denotes the sigmoid activation, and $c$ indexes feature channels. This dual branch formulation ensures both multimodal calibration (when $R>1$) and unimodal refinement (when $R=1$), effectively mapping heterogeneous sensor data to a unified feature space.

\subsection{Spatiotemporal Disentanglement Mechanism}

\begin{figure}[ht]
    \centering
    \includegraphics[width=0.8\linewidth]{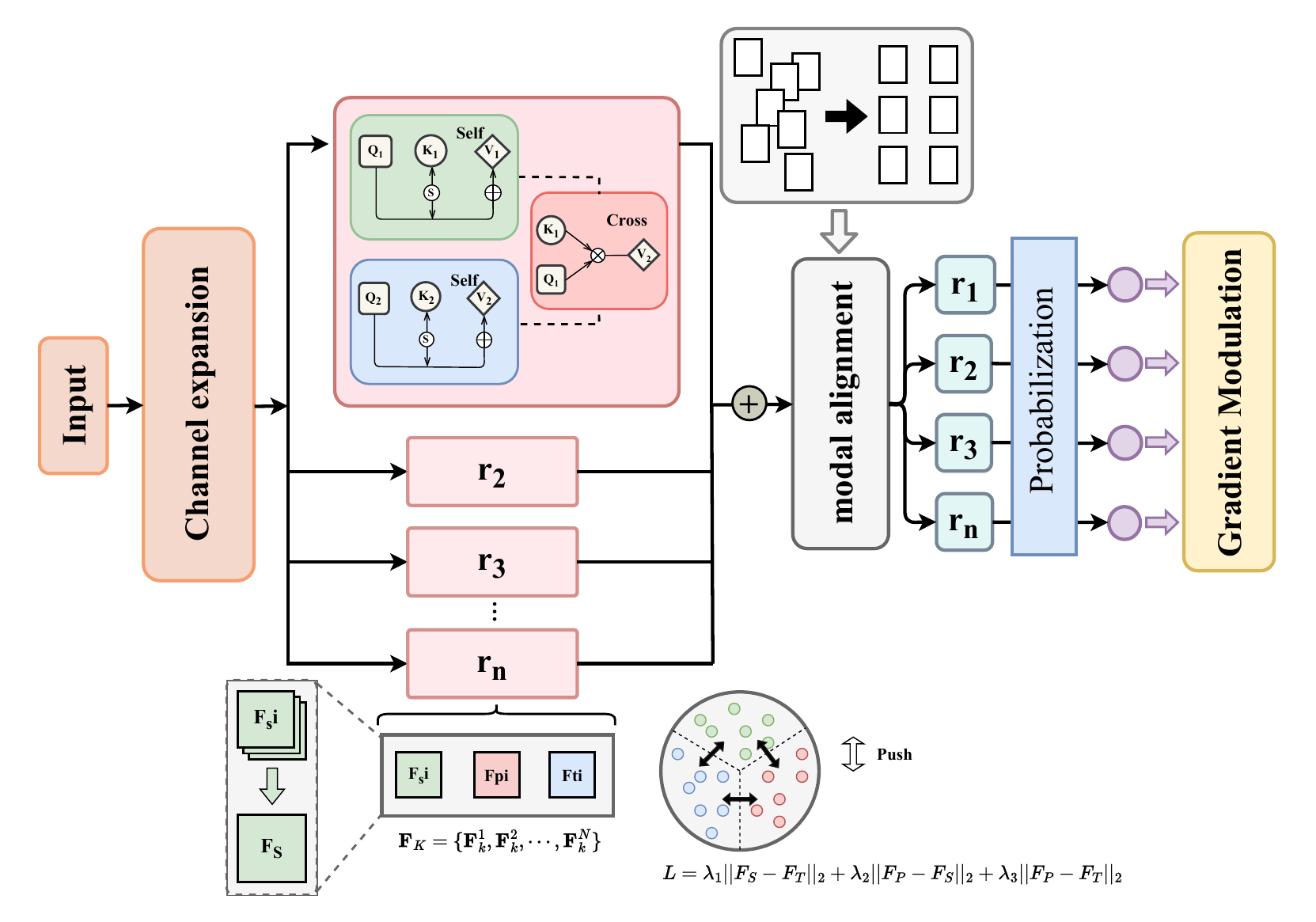}
    \caption{Detailed structure of the proposed CMD-HAR model}
    \label{fig:main method}
\end{figure}

Multisensor data often exhibit complex spatiotemporal characteristics. During the processing of such data, the mixing of spatial and temporal features often leads to redundancy and interference in information extraction. To address this issue, we designed a spatiotemporal disentanglement model, CMD-HAR, as shown in Figure\ref{fig:main method}. 

The weighted fusion $\mathbf{f}\in\mathbb{R}^{1\times S\times C/K}$
represented by the given cardinality group is used as input to the spatio-temporal disentanglement module to generate spatial self-attention features $\mathbf{F}_s^i$ and temporal self-attention features $\mathbf{F}_t^i$. Specifically, the embedding amount f is modeled using self-attention through the time disentangling module T(·) and the spatial disentangling module S(·), generating spatial features $\mathbf{F}_s^i$ and temporal features $\mathbf{F}_t^i$. At the same time, the public feature $\mathbf{F}_p^i$ is output through cross-attention modeling. $\mathrm{Attention}(Q,K,V)=\mathrm{softmax}(\frac{QK^T}{\sqrt{d_k}})V,$
where $Q=fW_Q,\quad K=fW_K,\quad V=fW_V,$
$W_{V}$, $W_{Q}$, and $W_{K}$ are learnable projection moments.
Finally, the three output features of each sensor group are combined to obtain a set of
$\mathbf{F}_{S}=\{\mathbf{F}_{s}^{1},\mathbf{F}_{s}^{2},\cdots,\mathbf{F}_{s}^{N}\},$
$\mathbf{F}_{T}=\{\mathbf{F}_{t}^{1},\mathbf{F}_{t}^{2},\cdots,\mathbf{F}_{t}^{N}\},$
$\mathbf{F}_{P}=\{\mathbf{F}_{p}^{1},\mathbf{F}_{p}^{2},\cdots,\mathbf{F}_{p}^{N}\}$.
In addition, we also propose the loss of cross-modal entanglement to minimize the differences between public representations and maximize the differences between independent representations. Formally, the definition of disentanglement loss is as follows: 
\vspace{-.5em}
\begin{equation}
L=\lambda_1{||\mathbf{F}_S-\mathbf{F}_T||_2}+\lambda_2{||\mathbf{F}_P-\mathbf{F}_S||_2}+\lambda_3{||\mathbf{F}_P-\mathbf{F}_T||_2},
\end{equation}

where $\lambda_1,\lambda_2,\lambda_3$ are adjustable parameters used to control the contribution of different loss items to the total loss. After completion of the spatiotemporal disentanglement, we perform a weighted fusion of the spatial feature set $\mathbf{F}_{S}$, the temporal feature set $\mathbf{F}_{T}$, and the public feature set $\mathbf{F}_{P}$. 

To address the issue of differences in the distribution of multi-modal features, we introduce a modality alignment module. This module dynamically generates attention weights to align and weigh multi-modal features. As a result, it enhances the model's focus on important modality information while suppressing interference from secondary modalities. During the forward propagation process, when the input feature $\mathbf{f}\in\mathbb{R}^{1\times S\times C/K}$
passes through the modality alignment module, the model performs first a channel grouping operation. We divide the number of channels \(C\) into groups \(M\), where \(M\) is equal to the number of modalities, so that each modal characteristic can be adjusted independently. For each group of channels, the following standard formula is used for calculation:

\vspace{-.5em}
\begin{equation}
\hat{\mathbf{f}}_{i} = \frac{\mathbf{f}_{i} - \mu}{\sqrt{\sigma^{2} + \epsilon}}\gamma + \beta,
\end{equation}

Specifically, \(\mu\) and \(\sigma^{2}\) are the mean and variance of the characteristic of all channels within the group, respectively; \(\epsilon\) is a constant used for numerical stability; \(\gamma\) and \(\beta\) are able to learn parameters for scaling and translation adjustments; \(\mathbf{f}_{i}\) represents the current value of the feature. The output feature $\hat{\mathbf{f}}_i\in\mathbb{R}^{1\times S\times C/K}$has the same shape as the input feature \(\mathbf{f}_{i}\), but its distribution is significantly normalized, forming a normalized multi-modal feature representation. 

After obtaining the grouped features, the module performs average pooling and max pooling operations on each group of features $\hat{\mathbf{f}}_{i}$ respectively, aiming to extract the global feature distribution and prominent feature information:
\vspace{-.5em}
\begin{equation}
P_{fused}=\frac{1}{H\times W}\sum_{i = 1}^{H}\sum_{j = 1}^{W}\mathbf{f}_{(:,i,j)}+\max_{\substack{i = 1,\ldots,H\\j = 1,\ldots,W}}\mathbf{f}_{(:,i,j)},
\end{equation}

The fused feature $P_{fused}\in R^{C}$ generates the attention weight $F_{2}$ through a fully connected two-layernetwork. Here, the first layer of the fully connected network $F_{1} = \phi(W_{1}P_{fused} + b_{1})$ is used to reduce the dimension of the characteristics, where \( \phi(x) = \max(0, x) \) represents the activation function ReLU. The second layer $F_{2}=W_{2}F_{1}+b_{2}$ restores the original dimension. $F_{2}$ is normalized by the Sigmoid activation function to generate the attention weight for each channel:
\vspace{-.5em}
\begin{equation}
A = \sigma(F_2) = \frac{1}{1 + e^{-F_2}},
\end{equation}
where $A\in R^{C}$ will be multiplied element-wise with the original features $\mathbf{f}_{(:,i,j)}$ to dynamically adjust the feature representation. Finally, the module performs residual connection through the CNN operator, combining the adjusted features $\hat{\mathbf{f}'}_{:,i,j} = A \odot \hat{\mathbf{f}}_{:,i,j}$ with the original input features $\mathbf{f}$, thus enhancing the model's robust modeling ability for multimodal information.

\subsection{Adaptive Modulation Gradient Balancing}  

Multimodal joint feature representation inherently suffers from modality heterogeneity, where dominant modalities tend to monopolize the optimization trajectory while suppressing gradient contributions from underperforming modalities. This phenomenon constrains the model's representational capacity and undermines multimodal synergy. To mitigate this fundamental challenge, we present Adaptive Modulation Gradient Balancing (AMGB), a dual-component framework that enforces equilibrium across modalities through dynamic gradient recalibration while enhancing generalization via stochastic exploration. As depicted in Figure \ref{framework}, our methodology operates through three coordinated mechanisms: gradient contribution monitoring, adaptive modulation, and stochastic enhancement.

\textbf{Gradient Contribution Monitoring: }
The AMGB framework begins with a continuous assessment of modality-specific gradient behaviors. During backpropagation, we quantify the magnitude of the instantaneous gradient$\|\nabla_{\theta}\mathcal{L}_m\|_2$ for each modality $m$, where $\mathcal{L}_m$ denotes the modality-specific loss component. This real-time monitoring enables detection of optimization dominance through comparative analysis of gradient contribution ratios:  \vspace{-.5em}
\begin{equation}
    \rho_m = \frac{\|\nabla_{\theta}\mathcal{L}_m\|_2}{\frac{1}{M-1}\sum_{n\neq m}\|\nabla_{\theta}\mathcal{L}_n\|_2},
\end{equation}
where $M$ represents total modalities. Ratios exceeding unity indicate dominant modalities requiring suppression, while subunity values signal under-optimized modalities needing enhancement.

\textbf{Adaptive Gradient Modulation: }
Building on the gradient dominance metric $\rho_m$, we devise a sigmoidal modulation operator with controlled saturation characteristics:  \vspace{-.5em}
\begin{equation}
    \gamma_m = \begin{cases} 
    1 - \tanh\left(\beta \cdot \text{ReLU}(\rho_m - 1)\right), & \rho_m > 1 \\
    1, & \text{otherwise}
    \end{cases}
\end{equation}  
The hyperbolic nonlinearity of the tangent ensures smooth transition between the modulation states, while the ReLU operator ($\text{ReLU}(x)=\max(0,x)$) activates suppression only for dominant modalities ($\rho_m>1$). This formulation preserves gradient directions while adaptively scaling magnitudes – critical for maintaining optimization stability.

\noindent
\textbf{Optimization Integration: }
We implement the modulation scheme through enhanced momentum estimation in the Adam optimizer. Let $g_t^m = \nabla_{\theta}\mathcal{L}_m$ denote the gradient for modality $m$ in step $t$. The modulated first-order momentum becomes:  \vspace{-.5em}
\begin{equation}
    m_t^m = \beta_1 m_{t-1}^m + (1-\beta_1)(\gamma_m \odot g_t^m),
\end{equation}  
where $\odot$ denotes element-wise multiplication. Concurrently, the second-order momentum $v_t^m$ remains unaltered to preserve gradient variance estimation. After bias correction, parameters update as:  \vspace{-.5em}
\begin{equation}
    \theta_{t+1}^m = \theta_t^m - \frac{\alpha \widehat{m}_t^m}{\sqrt{\widehat{v}_t^m} + \epsilon},
\end{equation}  
This selective modulation maintains Adam's convergence guarantees while enforcing inter-modal equilibrium.

\noindent
\textbf{Dynamic Stochastic Enhancement: }
To prevent equilibrium-induced local optima, we inject annealed Gaussian noise during parameter updates:  \vspace{-.1em}
\begin{equation}
    \theta_{t+1}^m \leftarrow \theta_{t+1}^m + \mathcal{N}(0, \sigma_t^2 I),
\end{equation}  
The noise variance $\sigma_t^2$ follows cosine annealing:  \vspace{-.1em}
\begin{equation}
    \sigma_t^2 = \sigma_{\text{max}}^2 \left(\frac{1}{2}\cos\left(\frac{\pi t}{T}\right) + \frac{1}{2}\right) + \sigma_{\text{min}}^2,
\end{equation}  
where $T$ denotes total training steps. This schedule promotes exploration in early training while converging to deterministic updates, balancing discovery and refinement.

The AMGB framework establishes a self-regulated optimization ecosystem where modalities compete through gradient-based signaling while cooperating via shared parameter space – a biologically inspired mechanism that surpasses conventional static balancing approaches.

\section{Design of Experiments}

\subsection{Dataset used in the experiments}

The proposed models are trained and evaluated using six different datasets. In order to ensure an objective evaluation of our methodology, several pertinent aspects of these used datasets have been outlined below. 

\textbf{WISDM} \cite{kwapisz2011activity}: Smartphone accelerometers recorded data at 20 Hz from 36 participants, capturing 18 activities such as walking, running, sitting, and climbing stairs. \textbf{UCI-HAR} \cite{anguita2013public}: This dataset was collected using waist-mounted smartphones with accelerometers and gyroscopes at 50 Hz, involving 30 participants and daily activities such as walking, running and sitting. \textbf{PAMAP2} \cite{reiss2012introducing}: With wearable sensors collecting data at 100 Hz, 9 participants performed 18 activities, including walking, running, and cycling. \textbf{USC-HAD }\cite{zhang2012usc}: Collected using wearable sensors at 50 Hz, this dataset includes 12 activities performed by 14 participants in their daily routines. \textbf{OPPORTUNITY} \cite{chavarriaga2013opportunity}: This dataset features 20 activities recorded with wearable and environmental sensors at multiple frequencies, designed for multimodal HAR research. \textbf{UniMiB-SHAR} \cite{micucci2017unimib}: Collected using smartphone accelerometers at 50 Hz, with 50 participants performing 30 activities (including daily behaviors and fall scenarios).

\begin{table}[!ht]
    \centering
    \scriptsize 
        \begin{tabular}{lcccccc}
            \midrule
            Datasets    & PAMAP2 & WISDM & OPPO. & UCI-HAR & UniMiB. & USC-HAD \\
            \midrule
            Sensor      & 40     & 3     & 72    & 9       & 3           & 6     \\
            Rate        & 33HZ   & 20HZ  & 30HZ  & 50HZ    & 50HZ        & 50HZ  \\
            Subject     & 9      & 29    & 12    & 30      & 50          & 14    \\
            Class       & 12     & 6     & 18    & 6       & 17          & 12    \\
            Window Size & 171    & 90    & 36      & 128     & 151         & 128   \\
            Tr/Va/Te    & 8:1:1  & 7:2:1 & 7:2:1 & 8:1:1   & 7:2:1       & 7:2:1 \\
            Batch size  & 128    & 512   & 256   & 512     & 64          & 64    \\
            Lr          & 0.001  & 0.001 & 0.001 & 0.001   & 0.001       & 0.001 \\
            \midrule
        \end{tabular}
    \caption{Dataset Details}
    \label{tab:plain}
\end{table}

To match the datasets for model testing, the datasets undergo a rigorous pre-processing process prior to formal training. Table \ref{tab:plain} shows the processing details and parameters of the datasets for CMD-HAR. 

\subsection{Evaluation Metrics}

To evaluate the performance of the proposed model for HAR, the following metrics were used. Accuracy is a measure of the predictability of the model. The F1 score is used to evaluate the balance between the accuracy and recall of the model when the classes are unbalanced. G-mean is used to evaluate the model's ability to identify minority classes in the case of class imbalance. For specific Metrics calculation formulas, please see the supplementary materials

\subsection{Experimental Environment and Design}

We utilized an NVIDIA V100 16GB GPU, combined with dual AMD EPYC processors (256 cores, 512 threads), along with other standard hardware configurations.

\section{Results and Discussion}

\subsection{Ablation Experiment}

We designed four ablation studies based on the CMD-HAR model, where we respectively removed the disentangling, gradient modulation, gradient expansion, and gradient alignment components. The experiments were conducted on the OPPORTUNITY and UCI-HAR datasets, with the results for the OPPORTUNITY dataset presented in Table \ref{tab:ablation oppo}, and the results for the UCI-HAR dataset shown as a confusion matrix in Figure 1 in our supplementary materials.

\begin{table}[!h]
    \centering
    \footnotesize
    \begin{tabular}{lccc}
        \midrule
        Model    & Accuracy  & F1-score & G-mean \\
        \midrule
        $w/o$ Disentangling  & 82.23   & 0.82  & 0.74\\
        $w/o$ Gradient Balancing   & 83.37   & 0.84  & 0.79 \\
        $w/o$ Gradient expansion & 80.75 & 0.81 & 0.73 \\
        $w/o$ Gradient alignment& 80.43 & 0.81 & 0.74 \\
        \textbf{CMD-HAR} & \textbf{86.23} & \textbf{0.85} & \textbf{0.91} \\
        \midrule
    \end{tabular}
    \caption{Ablation result on OPPORTUNITY.}
    \label{tab:ablation oppo}
\end{table}

After removing any module, the model's performance declined, with a particularly significant drop in G-mean, for instance, decreasing from 0.91 to around 0.75 on the OPPORTUNITY dataset. This indicates that the CMD-HAR model, through the synergistic effects of the disentangling module, gradient modulation, gradient expansion, and gradient alignment, can more accurately predict activity types that are challenging for other models to recognize. The removal of any of the gradient modulation, expansion, or alignment modules reduced the model's sensitivity to multimodal features, leading to a decrease in accuracy of approximately 5\% on the OPPORTUNITY dataset. Specifically, removing the gradient modulation module weakened the model's ability to handle class imbalance, removing the gradient expansion module reduced the learning of feature diversity, and removing the gradient alignment module disrupted the consistent integration of multimodal information.

The removal of the disentangling module also resulted in performance degradation, likely due to noise introduced by individual differences interfering with model learning. The disentangling module separates user-specific features from activity-related features, allowing the model to focus more on learning activity-related features, thereby improving generalization ability.

\begin{table}[!h]
    \centering
    \scriptsize
    \begin{tabular}{c|ccc|ccc|ccc|ccc}
    \toprule
        \multirow{2}{*}{\textbf{MODEL}} & \multicolumn{3}{c|}{\textbf{OPPO.}} & \multicolumn{3}{c|}{\textbf{SHAR}} & \multicolumn{3}{c|}{\textbf{WISDM}} & \multicolumn{3}{c}{\textbf{UCI-HAR}} \\ \cmidrule{2-13}
        & \textbf{acc} & \textbf{f1} & \textbf{g-m} & \textbf{acc} & \textbf{f1} & \textbf{g-m} & \textbf{acc} & \textbf{f1} & \textbf{g-m} & \textbf{acc} & \textbf{f1} & \textbf{g-m} \\ \midrule
        CNN\cite{zeng2014convolutional} & 85.09 & 0.85 & 0.79 & 75.92 & 0.76 & 0.65 & 96.50 & 0.94 & 0.94 & 93.18 & 0.93 & 0.93 \\ 
        LSTM\cite{xia2020lstm} & 85.91 & 0.86 & 0.79 & 84.25 & 0.84 & 0.76 & 99.60 & 0.99 & 0.99 & 88.12 & 0.88 & 0.88 \\ 
        CNN-GRU\cite{dua2021multi} & 74.86 & 0.73 & 0.72 & 83.07 & 0.83 & 0.74 & 81.11 & 0.81 & 0.80 & 90.80 & 0.91 & 0.91 \\ 
        LSTM-CNN\cite{xia2020lstm} & 81.08 & 0.81 & 0.75 & 87.84 & 0.88 & 0.81 & 98.97 & 0.98 & 0.98 & 89.04 & 0.89 & 0.89 \\ 
        Rev-Attention\cite{pramanik2023transformer} & 59.71 & 0.59 & 0.44 & 75.22 & 0.75 & 0.62 & 98.98 & 0.98 & 0.98 & 91.92 & 0.92 & 0.92 \\
        SE-Res2Net\cite{gao2019res2net} & 82.96 & 0.83 & 0.75 & 86.12 & 0.79 & 0.86 & 88.42 & 0.87 & 0.87 & 90.96 & 0.91 & 0.91 \\
        ResNext\cite{mekruksavanich2022deep} & 83.05 & 0.83 & 0.75 & 87.57 & 0.87 & 0.81 & 91.68 & 0.90 & 0.90 & 89.76 & 0.89 & 0.89 \\
        Gated-Res2Net\cite{yang2020gated} & 81.98 & 0.82 & 0.75 & 86.86 & 0.87 & 0.80 & 89.62 & 0.88 & 0.88 & 86.90 & 0.85 & 0.85 \\
        MAG-Res2Net\cite{liu2023mag} & 87.39 & 0.88 & 0.80 & 86.40 & 0.86 & 0.79 & 89.89 & 0.89 & 0.89 & 91.56 & 0.92 & 0.91 \\
        HAR-CE\cite{CE} & 83.05 & 0.83 & 0.74 & 86.43 & 0.86 & 0.79 & 91.89 & 0.91 & 0.91 & 91.35 & 0.91 & 0.91 \\
        ELK\cite{ELK} & 82.39 & 0.83 & 0.73 & 88.19 & 0.88 & 0.83 & 93.12 & 0.93 & 0.92 & 91.50 & 0.91 & 0.91 \\
        DanHAR\cite{DanHAR} & 86.16 & 0.86 & 0.77 & 86.00 & 0.86 & 0.78 & 94.35 & 0.94 & 0.93 & 91.44 & 0.91 & 0.92 \\
        RepMobile \cite{yu2024repmobile} & 72.47 & 0.72 & 0.69 & 83.66 & 0.83 & 0.74 & 90.30 & 0.91 & 0.90 & 92.68 & 0.93 & 0.91 \\
        SACNN-GA \cite{sarkar2023human} & 75.53 & 0.74 & 0.68 & 84.66 & 0.84 & 0.69 & 91.75 & 0.92 & 0.91 & 87.85 & 0.89 & 0.88 \\
        \textbf{Ours} & \textbf{86.23} & \textbf{0.85} & \textbf{0.91} & \textbf{87.88} & \textbf{0.82} & \textbf{0.96} & \textbf{98.14} & \textbf{0.97} & \textbf{0.99} & \textbf{96.53} & \textbf{0.91} & \textbf{0.95} \\
    \bottomrule
    \end{tabular}
    \caption{Comparison with Existing Work on the OPPORTUNITY, SHAR, WISDM, and UCI-HAR.}
    \label{comparison}
\end{table}

\subsection{Comparison with Existing Work}

We compared our model with various methods, including traditional CNN and LSTM network frameworks, as well as state-of-the-art models such as HAR-CE, ELK, DanHAR, RepMobile and SACNN-GA. To further evaluate our approach, we conducted comparisons with networks aligned with our strategy. For instance, SE-Res2Net, Gate-Res2Net, and MAGRes2Net leverage residual networks. We conducted tests on six datasets as outlined in Module 4.1, with the results of four datasets presented in Table \ref{comparison} and the remaining two datasets provided in the Supplementary Table.

It is noteworthy that the G-mean of CMD-HAR has improved by more than 0.2 across all datasets involving complex activities (OPPORTUNITY, PAMAP2, UniMiB-SHAR, USC-HAD), significantly outperforming other models. This indicates that CMD-HAR holds a substantial advantage in recognizing activity categories that are challenging for other models to identify, particularly complex activities. At the same time, CMD-HAR has also achieved outstanding accuracy levels of 98.14\% and 96.53\% on simpler activity datasets, such as WISDM and UCI-HAR, respectively. These results highlight the strong performance of our model on both complex and simple datasets, further demonstrating its effectiveness in addressing the challenges of multimodal complex activity recognition.

\begin{figure}[htbp]
    \centering
    \begin{minipage}[b]{0.37\textwidth}
        \includegraphics[width=\linewidth]{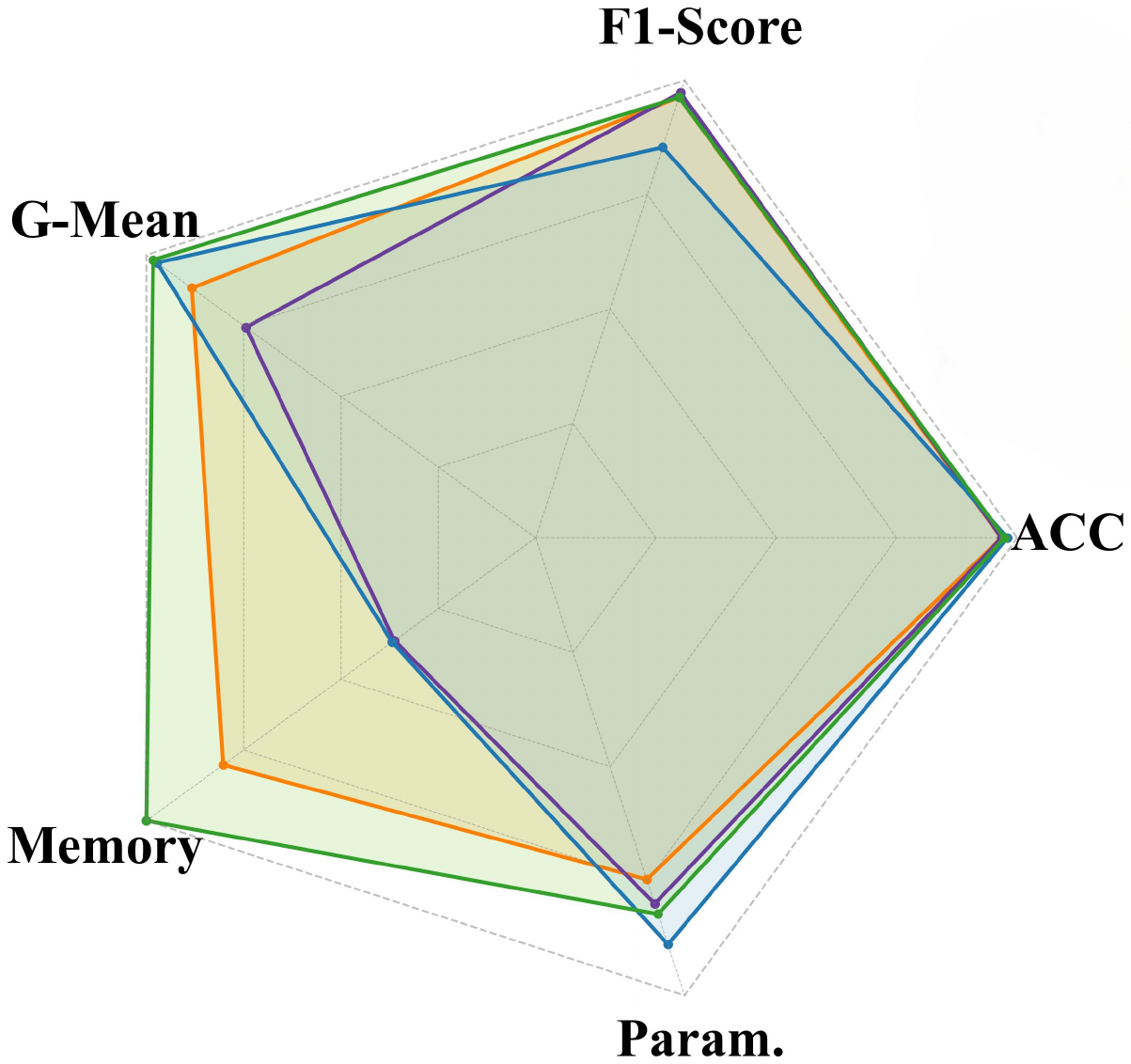}
        \textbf{\hspace*{2cm} (a) OPPO.}
        \label{fig:image1}
    \end{minipage}%
    \begin{minipage}[b]{0.48\textwidth}
        \includegraphics[width=\linewidth]{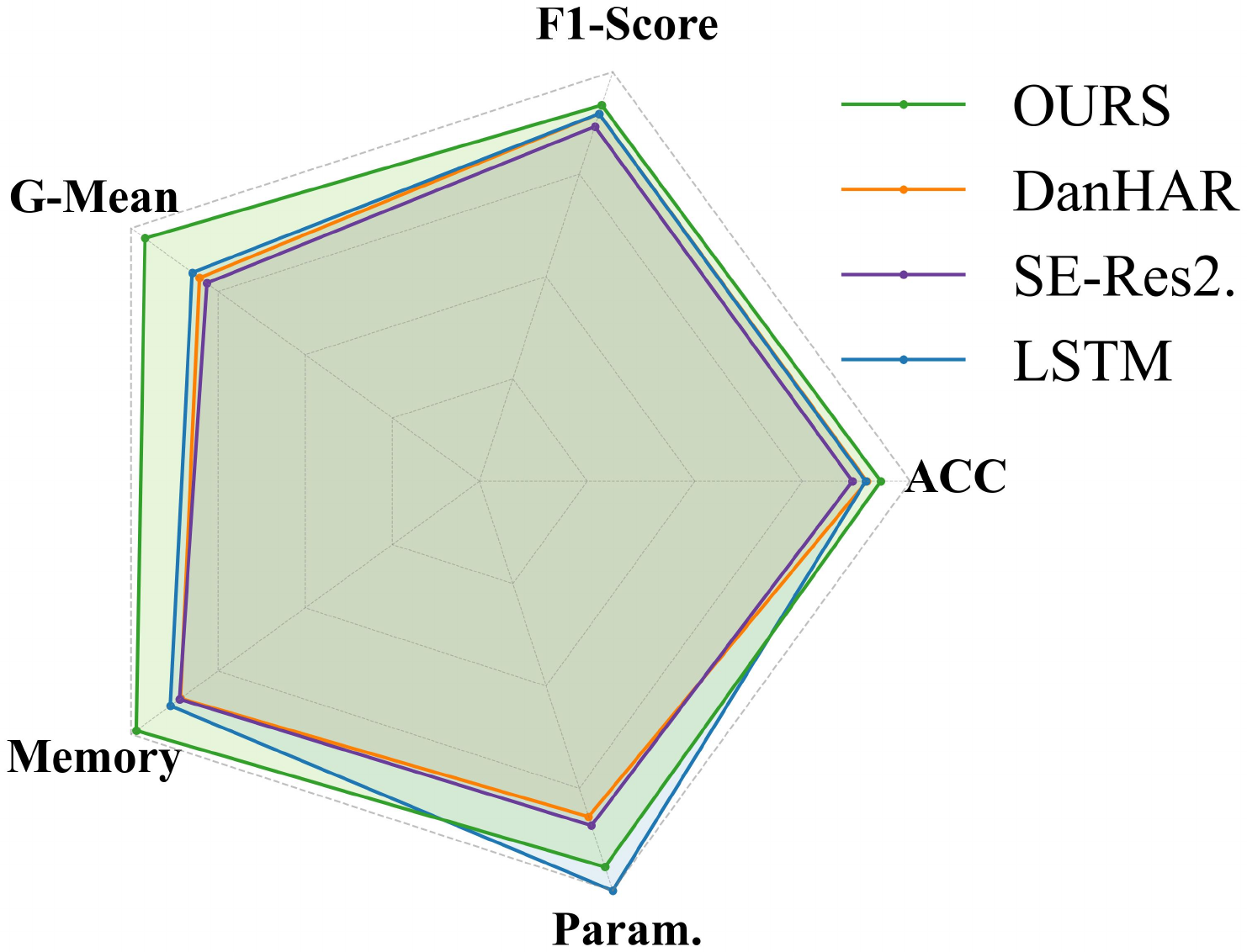}
        \textbf{\hspace*{2cm} (b) WISDM}
        \label{fig:image2}
    \end{minipage}
    \caption{Performance Comparison on OPPO. and WISDM.}
    \label{fig:images}
\end{figure}


\begin{table}[!h]
    \centering
    \scriptsize
    \begin{tabular}{c|ccc|ccc|ccc|ccc|ccc}
    \toprule
        \multirow{2}{*}{\textbf{UniMiB-SHAR}} & \multicolumn{3}{c|}{\textbf{ORIGINAL}} & \multicolumn{3}{c|}{\textbf{5}} & \multicolumn{3}{c|}{\textbf{10}} & \multicolumn{3}{c|}{\textbf{15}} & \multicolumn{3}{c}{\textbf{20}} \\ \cmidrule{2-16}
        & \textbf{Acc} & \textbf{F1} & \textbf{G-M} & \textbf{Acc} & \textbf{F1} & \textbf{G-M} & \textbf{Acc} & \textbf{F1} & \textbf{G-M} & \textbf{Acc} & \textbf{F1} & \textbf{G-M} & \textbf{Acc} & \textbf{F1} & \textbf{G-M} \\ \midrule
        CNN & 75.92 & 0.76 & 0.65 & 28.46 & 0.22 & 0.00 & 61.69 & 0.61 & 0.48 & 70.41 & 0.70 & 0.58 & 74.28 & 0.74 & 0.63 \\
        LSTM & 84.25 & 0.84 & 0.76 & 56.45 & 0.56 & 0.47 & 59.50 & 0.58 & 0.47 & 46.29 & 0.45 & 0.29 & 84.25 & 0.84 & 0.76 \\
        CNN-GRU & 83.07 & 0.83 & 0.74 & 24.86 & 0.18 & 0.00 & 60.09 & 0.57 & 0.00 & 74.67 & 0.74 & 0.67 & 77.00 & 0.75 & 0.00 \\
        LSTM-CNN & 87.84 & 0.88 & 0.81 & 67.71 & 0.67 & 0.50 & 84.48 & 0.84 & 0.76 & 86.71 & 0.87 & 0.79 & 86.83 & 0.87 & 0.79 \\
        Rev-Attention & 75.22 & 0.75 & 0.62 & 45.19 & 0.45 & 0.25 & 69.19 & 0.69 & 0.50 & 74.86 & 0.75 & 0.61 & 78.62 & 0.78 & 0.67 \\
        SE-Res2Net & 86.12 & 0.79 & 0.86 & 80.79 & 0.59 & 0.00 & 84.21 & 0.84 & 0.73 & 85.54 & 0.86 & 0.78 & 85.54 & 0.86 & 0.78 \\
        ResNext & 87.57 & 0.87 & 0.81 & 56.53 & 0.54 & 0.00 & 77.25 & 0.76 & 0.53 & 85.89 & 0.86 & 0.77 & 84.75 & 0.85 & 0.77 \\
        Gated-Res2Net & 86.86 & 0.87 & 0.80 & 50.86 & 0.47 & 0.00 & 81.78 & 0.81 & 0.67 & 87.14 & 0.87 & 0.80 & 86.71 & 0.87 & 0.80 \\
        MAG-Res2Net & 86.40 & 0.86 & 0.79 & 66.73 & 0.64 & 0.40 & 82.72 & 0.83 & 0.73 & 87.02 & 0.87 & 0.80 & 86.59 & 0.87 & 0.80 \\
        HAR-CE & 86.43 & 0.86 & 0.79 & 70.76 & 0.70 & 0.60 & 83.50 & 0.83 & 0.75 & 85.97 & 0.86 & 0.78 & 88.04 & 0.88 & 0.82 \\
        ELK & 88.19 & 0.88 & 0.83 & 25.92 & 0.19 & 0.00 & 54.38 & 0.49 & 0.26 & 84.17 & 0.84 & 0.74 & 88.66 & 0.89 & 0.83 \\
        DanHAR & 86.00 & 0.86 & 0.78 & 26.54 & 0.20 & 0.00 & 38.78 & 0.34 & 0.00 & 59.07 & 0.56 & 0.44 & 80.22 & 0.80 & 0.74 \\
        RepMobile & 83.66 & 0.83 & 0.74 & 25.24 & 0.18 & 0.00 & 39.47 & 0.38 & 0.00 & 46.58 & 0.46 & 0.38 & 76.89 & 0.76 & 0.71 \\
        SACNN-GA & 84.66 & 0.84 & 0.69 & 25.46 & 0.18 & 0.00 & 62.35 & 0.63 & 0.52 & 67.20 & 0.66 & 0.57 & 78.74 & 0.78 & 0.73 \\
        \textbf{Ours} & \textbf{87.88} & \textbf{0.82} & \textbf{0.96} & \textbf{54.53} & \textbf{0.40} & \textbf{0.44} & \textbf{78.81} & \textbf{0.67} & \textbf{0.72} & \textbf{86.12} & \textbf{0.79} & \textbf{0.90} & \textbf{87.72} & \textbf{0.82} & \textbf{0.94} \\
    \bottomrule
    \end{tabular}
    \caption{The results of the noise injection experiment on UniMiB-SHAR.}
    \label{UniMiB-SHAR}
\end{table}

Our research uses a radar chart Figure \ref{fig:images} to visually compare the performance of different metrics. Each axis represents a metric, and a larger area indicates better overall performance. From the chart, it is evident that our model demonstrates a substantial advantage on complex datasets. On simple datasets, our model still outperforms others in overall performance. Particularly in terms of memory usage, our model shows significant advantages across both types of datasets.

To enhance the model's robustness and generalization in real-world environments, we performed noise injection experiments on public datasets. Signal-to-Noise Ratio (SNR) is an indicator of the ratio between signal strength and noise intensity, commonly used to describe signal quality. By setting different SNR values (5, 10, 15, 20), we simulated noise disturbances that may occur in practical applications, assessing the model's performance under different noise conditions. We applied noise to the six datasets mentioned in Section 4.1 and compared the results with those from the original datasets. The result for UniMiB-SHAR is presented in Table \ref{UniMiB-SHAR}, and the remaining results are in the appendix.
\begin{figure}[ht]
    \centering
    \includegraphics[width=1\linewidth]{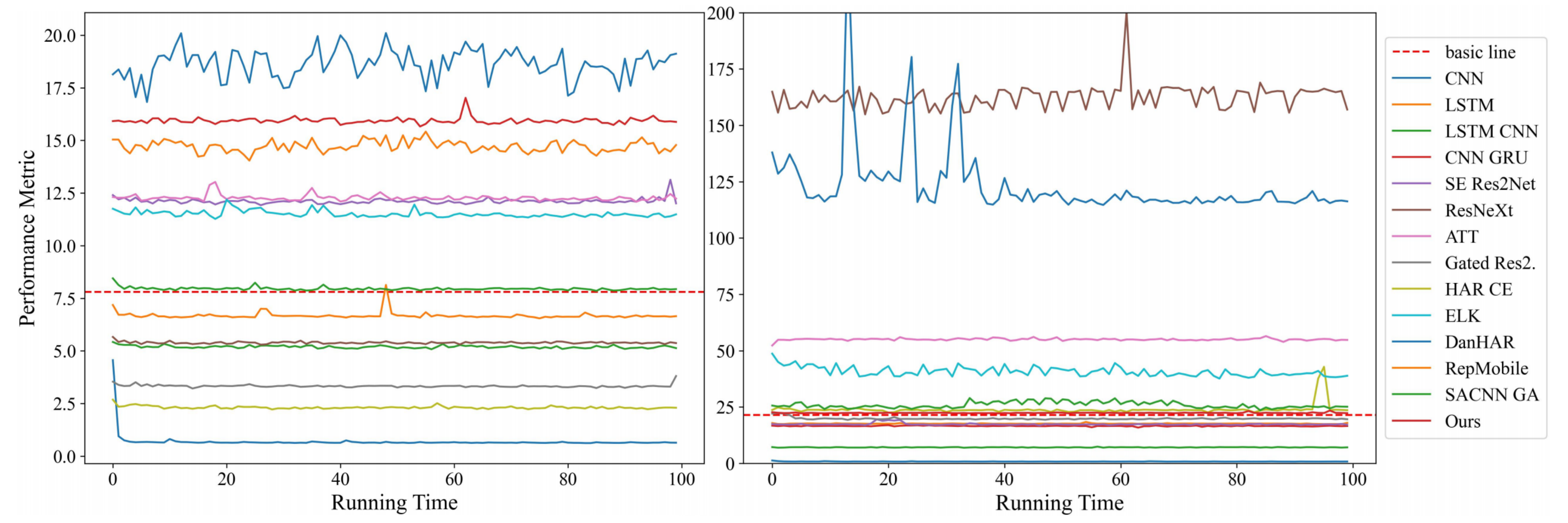}
    \caption{Inference delay of each model. (a) WISDM, (b) PAMAP2}
    \label{fig:3}
\end{figure}

The experimental results demonstrate that our model outperforms others under all four SNR conditions, particularly excelling on more complex datasets such as PAMAP2 and UniMiB-SHAR. Even under low SNR conditions, the model maintains high accuracy. 

\subsection{Actual Deployment}

We conducted four complexity evaluation experiments on the OPPORTUNITY, WISDM, and PAMAP2 datasets—model parameter size, memory consumption, inference delay, and single-segment inference delay. 
The evaluation was conducted on a Raspberry Pi 5 with a quad-core Cortex-A76  64-bit SoC clocked at 2.4 GHz, running PyTorch version 2.2.2.

\begin{table}[!h]
    \centering
    \scriptsize
    \begin{tabular}{c|cc|cc}
    \toprule
        \multirow{2}{*}{\textbf{MODEL}} & \multicolumn{2}{c|}{\textbf{WISDM}} & \multicolumn{2}{c}{\textbf{OPPO.}} \\ \cmidrule{2-5}
        & \textbf{Memory} & \textbf{Param.} & \textbf{Memory} & \textbf{Param.}\\ \midrule
        CNN & 766.23 & 1.04E+05 & 744.52 & 1.21E+05 \\
        LSTM & 759.80 & 1.65E+05 & 748.44 & 3.85E+05 \\
        CNN-GRU & 787.05 & 4.38E+06 & 773.17 & 4.39E+06 \\
        LSTM-CNN & 823.00 & 2.12E+06 & 788.69 & 2.13E+06 \\
        SE-Res2Net & 787.67 & 1.60E+06 & 772.41 & 1.61E+06 \\
        ResNeXt & 889.78 & 2.20E+07 & 849.25 & 2.21E+07 \\
        Rev-Att & 862.03 & 4.33E+05 & 801.97 & 3.80E+05 \\
        Gated-Res2Net & 773.84 & 1.60E+06 & 769.53 & 1.61E+06 \\
        HAR-CE & 788.73 & 4.17E+05 & 779.84 & 3.49E+06 \\
        ELK & 782.70 & 2.41E+05 & 751.84 & 7.57E+05 \\
        DanHAR & 790.66 & 2.35E+06 & 814.09 & 4.46E+06 \\
        RepMobile & 658.28 & 5.42E+05 & 703.48 & 5.48E+05\\
        SACNN-GA & 809.86 & 2.51E+06 & 659.36 & 2.59E+06\\
        \textbf{Ours} & \textbf{718.50} & \textbf{3.47E+05} & \textbf{751.30} & \textbf{1.09E+06} \\
    \bottomrule
    \end{tabular}
    \caption{Complexity analysis.}
    \label{tab:Complexity analysis}
\end{table}

To ensure test stability, we ran 100 iterations for all models, evaluating metrics such as running time and memory usage. Results for running the models on the full dataset were not recorded, as most models failed to complete the test. Additionally, we believe such a test has limited practical value without proper memory management adaptation. Table \ref{tab:Complexity analysis} shows the average results from ten rounds of 100 tests on the WISDM and OPPORTUNITY datasets, demonstrating that our method has lower memory consumption and fewer parameters compared to most state-of-the-art models. And the basic line in Table \ref{tab:Complexity analysis} means the approximate result of our model.

To evaluate efficiency of the model, we conducted inference delay tests, measuring the time taken to infer a single sample and 100 consecutive samples. We followed Zhang’s inference delay testing method, using a sliding window approach to evaluate sensor data \cite{NNU2023}, selecting 10-second action clips with a sliding window step of 95 \% of the window length (500 ms). Due to improved device performance, most models easily met the required standard. Considering that the WISDM dataset typically uses 4-second segments with a 200-millisecond step, we applied this method to each data set to ensure the model processes the next segment in time.

The experimental results are shown in Figure \ref{fig:3}. Some higher delay results are not shown in the figures. In the WISDM dataset, most models are concentrated in the lower time intervals. In the PAMAP2 dataset, there is a clear performance gap, with the basic model maintaining lower times, and lightweight models such as HAR-CE, ELK, and ours staying within 25 ms, while models like ResNeXt exceed the time limit.

\section{Conclusion}

In this study, we explored the challenges in sensor-based HAR, such as multimodal data mixing, and proposed the CMD-HAR method, which includes a spatiotemporal attention-based strategy, cross-modal disentangling, and cross-modal gradient modulation. We also developed a portable deployment system. Experiments on multiple datasets showed that our approach significantly improved recognition accuracy, universality, efficiency, and reduced resource consumption. Despite its added complexity, our method outperformed previous approaches. Our research contributes to the development of HAR. In the future, we will explore advanced techniques and optimize the model structure through network models to reduce complexity and enhance performance in complex scenarios.

\bibliographystyle{elsarticle-num}

\bibliography{refs}

\end{document}